\documentclass[runningheads]{llncs}
\usepackage{amsfonts}       % blackboard math symbols
\usepackage{amsmath}

\usepackage{psfrag}
\usepackage{graphicx}
\usepackage[numbers]{natbib}
\begin{document}
\title{Highway State Gating for Recurrent Highway Networks: improving information flow through time}
\titlerunning{Highway State Gating for RHN: improving information flow through time}

\author{Ron Shoham \and
Haim Permuter }
\authorrunning{R. Shoham, H. Permuter}
\institute{Ben-Gurion University, Beer-Sheva 8410501, Israel\\
\email{ronshoh@post.bgu.ac.il}\\
\email{haimp@bgu.ac.il}\\
}
\maketitle              % typeset the header of the contribution
\begin{abstract}
Recurrent Neural Networks (RNNs) play a major role in the field of sequential learning, and have outperformed traditional algorithms on many benchmarks. Training deep RNNs still remains a challenge, and most of the state-of-the-art models are structured with a transition depth of 2-4 layers. Recurrent Highway Networks (RHNs) were introduced in order to tackle this issue. These have achieved state-of-the-art performance on a few benchmarks using a depth of 10 layers. However, the performance of this architecture suffers from a bottleneck, and ceases to improve when an attempt is made to add more layers. In this work, we analyze the causes for this, and postulate that the main source is the way that the information flows through time. We introduce a novel and simple variation for the RHN cell, called Highway State Gating (HSG), which allows adding more layers, while continuing to improve performance. By using a gating mechanism for the state, we allow the net to "choose" whether to pass information directly through time, or to gate it. This mechanism also allows the gradient to back-propagate directly through time and, therefore, results in a slightly faster convergence. We use the Penn Treebank (PTB) dataset as a platform for empirical proof of concept. Empirical results show that the improvement due to Highway State Gating is for all depths, and as the depth increases, the improvement also increases.  
\end{abstract}
\section{Introduction}
Training very deep neural networks has become very common in the last few years. Both theoretical and empirical evidence points to the fact that deeper networks can represent more efficiently specific functions (\citet{bengio2007scaling, bianchini2014complexity}). Some commonly used architectures for deep feed-forward networks are Resnet\cite{Resnets}, Highway Networks \cite{srivastava2015highway} and Dense-Net\cite{huang2017densely}. These architectures can be structured with tens, and sometimes even hundreds of layers. Unfortunately, training a very deep Recurrent Neural Network (RNN) still remains a challenge.

\citet{RHN} introduced the Recurrent Highway Network (RHN) in order to address this issue. Its main difference from previous deep RNN architectures, was incorporating Highway layers inside the recurrent transition. By using a transition depth of 10 Highway layers, RHN managed to achieve state-of-the-art results on several benchmarks of word and character prediction. However, increasing the transition depth of a similar RHN, does not improve the results significantly. 

In this paper, we first analyze the reasons for this phenomena. Based on the results of our analysis, we suggest a simple solution which adds a non-significant number of parameters. This variant is called a \textit{Highway State Gating} cell or a \textit{HSG}. By using the HSG mechanism, the new state is generated by a weighted combination of the previous state and the output of the RHN cell. The main idea behind the HSG cell is to provide a fast route for the information to flow through time. That way, we also provide a shorter path for the back-propagation through time (BPTT). This enables the use of a deeper transition depth, together with significant performance improvement on a widely used benchmark.

\section{Related Work}
Gated-Recurrent-Units (GRUs) \cite{cho2014enc} were suggested in order to reduce the number of parameters of the traditional and commonly used Long-Short-Term-Memory (LSTM) cell (\citet{HOCHREITER}). Similarly to HSG, in GRUs the new state is a weighted sum of the previous state and a non-linear transition of the current input and the previous state. The main difference is that the transition is of a depth of a single layer and, therefore, less robust.

\citet{kim2017residual} introduced a different variant of the LSTM cell which is inspired by Resnet\cite{Resnets}. They proposed adding to the LSTM cell a residual connection from its input to the reset gate projection output. By that they allowed another route for the information to flow directly through. They managed to train a net of 10 residual LSTM layers which outperformed other architectures. In their work, they focused on the way that the information passes through layers in the feed-forward manner, and not on the way it passes through time.

\citet{wang2016recurrent} used residual connections in time. In their work they talked about the way information passes through time. They managed to improve performance on some benchmarks, while reducing the number of parameters. The difference is that they needed to work with a fixed residual length that is a hyper-parameter. Also, their work focused on cells with a one layer transition depth.

Another article, relating to Zoneout regularization (\citet{krueger2016zoneout}) also relates to information flow through time. The authors introduced a new regularization method for RNNs, where the idea is very similar to dropout\cite{dropout}. The difference is that the dropped neurons in the state vectors get their values in the former time-step, instead of being zeroed. They mentioned that one of the benefits of this method is that the BPTT skips a time-step on its path back through time. In our work, there is a direct (weighted) connection between the current state and the former one, which is used similarly both for training and inference.

Another relevant issue is the \textit{slowness} regularizers (\citet{hinton1990connectionist, foldiak1991learning, luciw2012low, jonschkowski2015learning, merity2017revisiting}) which add a penalty for large changes in state through time. In our work we do not add such a penalty, but we allow a direct route for the state to pass through time-steps, and therefore we 'encourage' the state not to change when it is not needed.

\section{Revisiting Vanilla Recurrent Highway Networks}
\begin{figure}[!ht]
  \centering
  \psfrag{g}[c][][.8]{$l=1$}
  \psfrag{h}[c][][.8]{$l=2$}
  \psfrag{j}[c][][.8]{$l=L$}
  \psfrag{c}[c][][.8]{$x^{[t]}$}
  \psfrag{f}[c][][.8]{$c_1^{[t]}$}
  \psfrag{e}[c][][.8]{$h_1^{[t]}$}
  \psfrag{d}[c][][.75]{$t_1^{[t]}$}
  \psfrag{k}[c][][.8]{$s_1^{[t]}$}
  \psfrag{z}[c][][.8]{$s_{L-1}^{[t]}$}
  \psfrag{a}[c][][.8]{$s_L^{[t-1]}$}
  \psfrag{b}[c][][.8]{$s_L^{[t]}$}
  \includegraphics[scale=0.5]{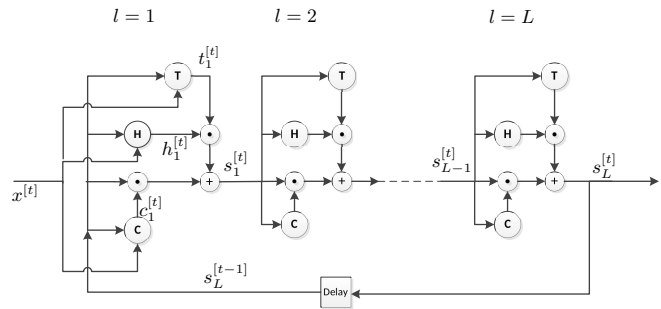}
  \caption{\textbf{Schematic showing RHN cell computation.} The Feed-Forward route goes from bottom to top through $L$ stacked Highway layers. On the right side there is the memory unit, followed by the recurrent connection.}
  \label{RHN_fig}
\end{figure}

Let $L$ be the transition depth of the RHN cell, and $x^{[t]}\in \mathbb{R}^m$ be the cell's input at time $t$. Let $W_{H,T,C}\in \mathbb{R}^{n\times m}$ and $R_{H_l,T_l,C_l}\in \mathbb{R}^{n\times n}$ represent the weight matrices of $H$ nonlinear transforms and the $T$ and $C$ gates at layer $l\in \{1,\dots, L\}$. The biases are denoted by $b_{H_l,T_l,C_l} \in \mathbb{R}^n$, and let $s_l^{[t]}$ denote the intermediate output at layer $l$ at time $t$, with $s_0^{[t]} = s_L^{[t-1]}$. The gates $T$ and $C$ utilize a sigmoid ($\sigma$) non-linearity and "$\cdot$" denotes element-wise multiplication.
An RHN layer is described by 
\begin{align}
    s_l^{[t]} &= h_l^{[t]}\cdot t_l^{[t]} + s_{l-1}^{[t]}\cdot c_l^{[t]},
\end{align}
where 
\begin{align}
    h_l^{[t]} &= \tanh(W_Hx^{[t]}\mathbb{I}_{\{l=1\}} + R_{H_l}s_{l-1}^{[t]} + b_{H_l}),\\
    t_l^{[t]} &= \quad\,\: \sigma(W_Tx^{[t]}\mathbb{I}_{\{l=1\}} + R_{T_l}s_{l-1}^{[t]} + b_{T_l}),\\
    c_l^{[t]} &= \quad\,\: \sigma(W_Cx^{[t]}\mathbb{I}_{\{l=1\}} + R_{C_l}s_{l-1}^{[t]} + b_{C_l}),
\end{align}
and $\mathbb{I}$ is the indicator function. A very common variant for this is coupling gate $C$ to gate $T$, i.e. $C=1-T$. Figure \ref{RHN_fig} illustrates the RHN cell.

\begin{figure}[!ht]
\centering
  \psfrag{C}[r][][.8]{layer 1}
  \psfrag{B}[r][][.8]{layer 2}
  \psfrag{A}[r][][.8]{layer L}
  \psfrag{D}[c][][.8]{$t$}
  \psfrag{Z}[c][][.8]{$t+1$}
  \psfrag{F}[c][][.8]{$t+T$}
  \psfrag{G}[l][][.8]{$x_t$}
  \psfrag{H}[l][][.8]{$x_{t+1}$}
  \psfrag{I}[l][][.8]{$x_{t+T}$}
   \includegraphics[scale=0.45]{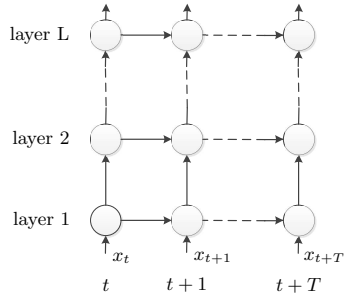}
   \caption{The figure illustrates an unfolded RNN with $L$ stacked layers. Here the path length from time $t$ to time $t+T$ is $L+T-1$.}
   \label{stacked_rnn} 
\end{figure}

\begin{figure}[!ht]
\centering
  \psfrag{A}[lb][][.8]{layer 1}
  \psfrag{B}[lb][][.8]{layer L}
  \psfrag{J}[c][][.8]{$t$}
  \psfrag{K}[l][][.8]{$t+1$}
  \psfrag{L}[l][][.8]{$t+T$}
  \psfrag{F}[b][][.8]{$s_L^{[t-1]}$}
  \psfrag{C}[bl][][.8]{$s_L^{[t]}$}
  \psfrag{D}[bl][][.8]{$s_L^{[t+1]}$}
  \psfrag{E}[bl][][.8]{$s_L^{[t+T]}$}
  \psfrag{G}[c][][.8]{$x_t$}
  \psfrag{H}[l][][.8]{$x_{t+1}$}
  \psfrag{I}[l][][.8]{$x_{t+T}$}
   \includegraphics[width=1\linewidth]{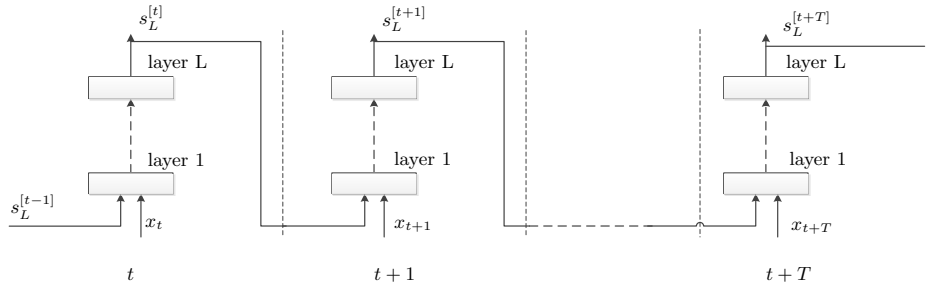}
   \caption{The figure illustrates an unfolded RHN with $L$ layers. Here the path length from time $t$ to time $t+T$ is $L\times T$.}
   \label{rhn_unfold}
\end{figure}

According to \citet{RHN}, one of the main advantages of using deep RHN instead of stacked RNNs, is the path length. While the path length of $L$ stacked RNNs from time $t$ to time $t+T$ is $L+T-1$ (figure \ref{stacked_rnn}), the path length of a RHN of depth $L$ is $L \times T$ (figure \ref{rhn_unfold}). The high recurrence depth can add significantly higher modeling power.

% \caption{The figures illustrates a comparison between the information path of (a) $L$ stacked RNNs, and (b) a RHN of depth $L$. The path length of stacked RNNs is $L+T-1$, whereas for RHN its $L\times T$.}

We believe that its power might, sometimes, also be its weakness. Let us examine a case where information that is relevant for a large number of time steps is given at time $t$; for example in stocks forecasting, where we expect a sharp movement to occur in the next few time steps. We would like the state to remain the same until the event happens (unless any dramatic event changes the forecast). In this case, we probably prefer the net state to remain stable without dramatic changes. However, when using a deep RHN, the information must pass through many layers, and that might cause an unwanted change of the state. For example, with a RHN of depth $30$, the input state at time $t$ has to pass $300$ layers in order to propagate $10$ time steps. To the best of our knowledge, there is no use of a feed-forward Highway Network of this depth in any field. This fact also affects the vanishing gradient issue using BPTT. The fact that the gradient needs to back-propagate through hundreds of layers causes it to vanish and not be effective. The empirical results support our assumption, and it seems like a performance bottleneck occurs when we use deeper nets.

\section{Highway State Gate in time}
\begin{figure}[!ht]
\centering
  \psfrag{A}[c][][.85]{$\hat{s}^{[t-1]}$}
  \psfrag{B}[c][][.85]{$s_{L}^{[t]}$}
  \psfrag{C}[c][][.85]{$1-g$}
  \psfrag{D}[c][][.85]{$g$}
  \psfrag{E}[b][][.85]{$\hat{s}^{[t]}$}
   \includegraphics[width=1\linewidth]{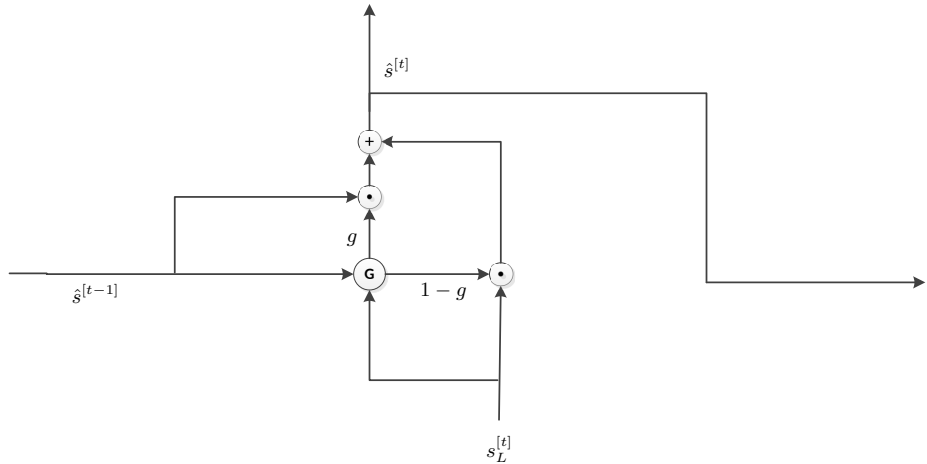}
   \caption{The figure illustrates a zoom into the HSG cell.}
   \label{HSG} 
\end{figure}

\begin{figure}[!ht]
\centering
  \psfrag{A}[l][][.8]{layer 1}
  \psfrag{B}[l][][.8]{layer L}
  \psfrag{C}[lt][][.8]{$s_{L}^{[t]}$}
  \psfrag{D}[lt][][.8]{$s_{L}^{[t+1]}$}
  \psfrag{E}[lt][][.8]{$s_{L}^{[t+L]}$}
  \psfrag{J}[c][][.8]{$t$}
  \psfrag{K}[c][][.8]{$t+1$}
  \psfrag{L}[c][][.8]{$t+T$}
  \psfrag{Z}[c][][.7]{$\;\;$ \textbf{HSG}}
  \psfrag{M}[c][][.8]{$\hat{s}^{[t-1]}$}
  \psfrag{F}[b][][.8]{$\hat{s}^{[t-1]}$}
  \psfrag{S}[l][][.8]{$\hat{s}^{[t]}$}
  \psfrag{R}[lb][][.8]{$\hat{s}^{[t+1]}$}
  \psfrag{Q}[lb][][.8]{$\hat{s}^{[t+T]}$}
  \psfrag{G}[c][][.8]{$x_t$}
  \psfrag{H}[l][][.8]{$x_{t+1}$}
  \psfrag{I}[l][][.8]{$x_{t+T}$}
   \includegraphics[width=1\linewidth]{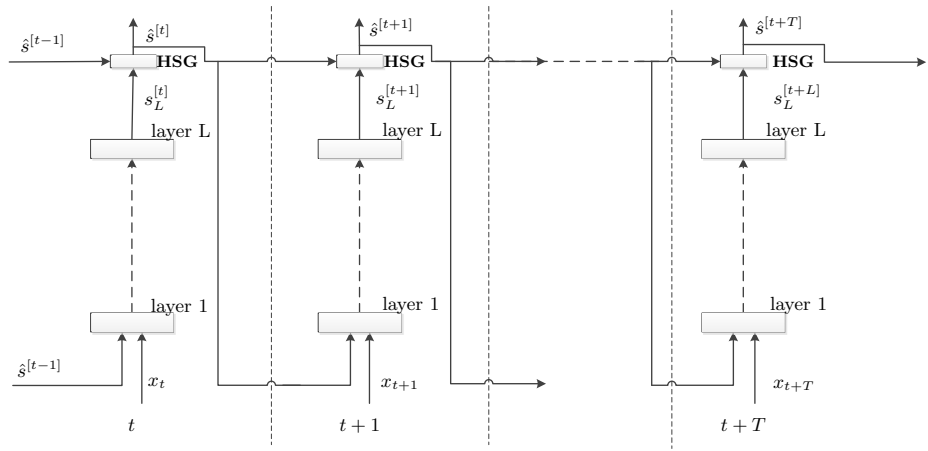}
   \caption{A macro scheme of an unfolded RHN with HSG cell. The state feeds both the RHN and the next time-step HSG cell.}
   \label{RHN_HSG_unfold}

% \caption{The figures illustrates the functionality of the HSG cell: (a) is a zoom into the cell structure; (b) is a macro scheme of an unfolded RHN with a HSG cell.}
\end{figure}

We suggest a simple, yet efficient, solution for the depth-performance bottleneck issue. Let $W_{R,F}\in \mathbb{R}^{n\times n}$ represent the weight matrices, and let $b_G\in \mathbb{R}^n$ be a bias vector. Let $s_{L}^{[t]}$ represent the output of the RHN cell at time $t$. $\hat{s}^{[t]}$ is the output of the HSG cell at time $t$. The HSG cell is described by
\begin{align}
    \hat{s}^{[t]} = g\cdot \hat{s}^{[t-1]} + (1-g)\cdot s_L^{[t]},
\end{align}
where
\begin{align}
    g^{[t]} &= \sigma(W_R\hat{s}^{[t-1]} + W_Fs_{L}^{[t]} + b_G).
\end{align}
A scheme of the HSG cell and an unfolded RHN with HSG is depicted in figure \ref{HSG} and figure \ref{RHN_HSG_unfold}, respectively. The direct outcome of adding an HSG cell is giving the information an alternative and fast route to flow through time.

Since gate $g$ utilizes a Sigmoid, its values are in the range $[0,1]$. When $g=0$, i.e. HSG is closed, $\hat{s}^{[t]} = s_L^{[t]}$. When $G=1$, i.e. the gate is opened, $\hat{s}^{[t]} = \hat{s}^{[t-1]}$. In the first case, the net functions as a vanilla RHN. In this case the information from the former state passes only through the functionality of the RHN. This means that the functionality of a regular RHN can be achieved easily even after stacking the HSG layer. 
\par One of the strengths of this architecture is that each state neuron has its own stand-alone gate. This means that some of the neurons can pass information easily through many time-steps, whereas other neurons learn short time dependencies.
\par Now let us examine the example we mentioned above, when using RHN with the HSG cell. The net depth is $30$, and a state needs to propagate $10$ time-steps. In this case, the state has multiple routes to propagate through. The propagation lengths are now $10 + 30j$, with $j\in \{0,1\dots 10\}$. This means that the information has multiple routes, and even if we use a really deep net, it still has a short path to flow through. For this reason, we expect our variant to enable training deeper RHNs more efficiently. The results below support our claim.

\section{Results}
Our experiments study the benefit of adding depth to a RHN with and without stacking HSG cells at its output. We conducted our experiments on the Penn Treebank (PTB) benchmark.
\par \textbf{PTB:} The Penn Treebank\footnote[1]{http://www.fit.vutbr.cz/˜imikolov/rnnlm/simple-examples.tgz}, presented by \citet{Marcus}, is a well known data set for experiments in the field of language modeling. The goal is predicting the next word at each time step, based on the past. Its vocabulary size is $10$k unique words. All words that are not in the vocabulary are labeled to a single token. The database is structured of $929$k training words, $73$k validation words, and $82$k test words.
\par
\begin{figure}[!ht]
  \centering
  \includegraphics[scale=0.45]{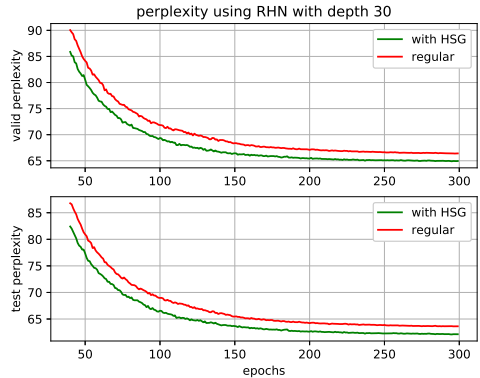}
  \caption{Comparison of the learning curve between RHN with (green) and without (red) HSG cell. The upper and the lower graphs show the perplexity on the validation and test sets respectively.}
  \label{learning_curve}
\end{figure}

 We used a hidden size of 830, similarly to that used by \citet{RHN}. For regularization, we use variational dropout \cite{VariationalDropout}, and L2 weight decay. The learning rate exponentially decreased at each epoch. An initial bias of $-2.5$ was used for both the RHN and the HSG gates. That way, the gates are closed at the beginning of training. We tried RHN depths from $\{10,20,30,40\}$. Results are shown in table \ref{res_table}. It can be well seen from the results that a performance bottleneck occurs when adding more layers to the vanilla RHN. However, adding more layers to the RHN network with the HSG cell results in a steady improvement. Figure \ref{learning_curve} also illustrates the difference between both architectures during training. It can be seen that not only does the net with HSG achieve better results, it also converges a bit faster than the vanilla one. Another interesting aspect is the histogram of the gate values of the HSG cell in figure \ref{val_hist}. It can be seen that most of the gates are usually closed (small valued). However, in a significant number of cases the gates open, which means that the model passes a very similar state to the next time step.
\begin{table}[!ht] 
  \caption{Single RHN model test and validation perplexity of the PTB dataset}
  \centering
  \begin{tabular}{|l|ll|ll|}
    \hline
     &
      \textbf{Validation set} & &
      \textbf{Test set} &\\
      \hline
      \textbf{RHN} & {\textbf{with HSG}} & {\textbf{w/o HSG}} & {\textbf{with HSG}} & {\textbf{w/o HSG}} \\
      \hline
    \textbf{depth=10}  & \textbf{67.5} & \textbf{67.9} & \textbf{65.0} & \textbf{65.4}\\
    \textbf{depth=20}  & \textbf{65.6} & \textbf{66.4} & \textbf{62.9} & \textbf{63.2}\\
    \textbf{depth=30}  & \textbf{64.8} & \textbf{66.4} & \textbf{62.0} & \textbf{63.4}\\
    \textbf{depth=40}  & \textbf{64.7} & \textbf{66.7} & \textbf{61.7} & \textbf{63.6}\\
    \hline
  \end{tabular}
  \vspace{1ex}\\
     *Note that the HSG is more significant as the depth of the RHN increases.
  \label{res_table}
\end{table}

\begin{figure}[!ht]
  \centering
  \includegraphics[scale=0.55]{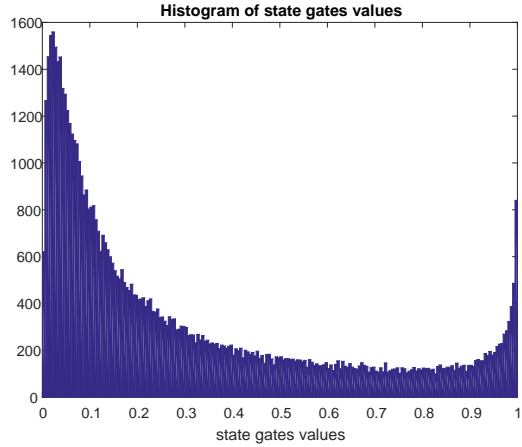}
  \caption{Histogram of HSG cell gates values. The values were drawn from a trained RHN of depth $30$, with a hidden size of $830$. There are $66400$ values from a $80$ random time steps. The gates utilize a Sigmoid function and, therefore, the values are in the range of $[0,1]$. We see that in most of the cases the gate values are relatively low, which means that the state gates are closed, and the new state is generated in a feed-forward manner. However, for a substantial number of times, the values are high, which means that the information flows directly through time.}
  \label{val_hist}
\end{figure}

\begin{figure}[!ht]
  \centering
  \includegraphics[scale=0.35]{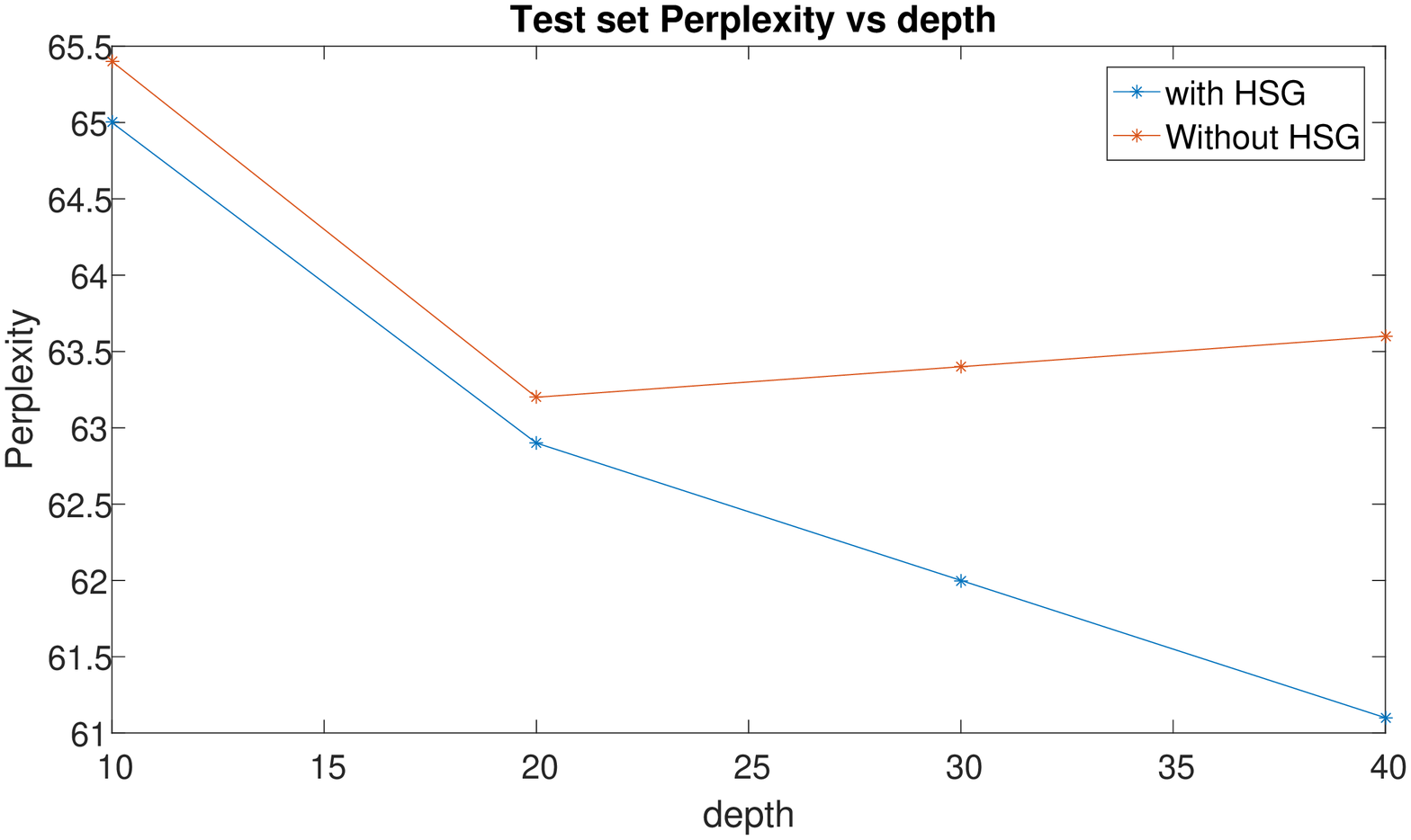}
  \caption{Graph of Perplexity vs depth of the RHN over the test set with (blue) and without(orange) HSG cell. This figure illustrates the depth-performance bottleneck phenomena. It can be seen that by the depth of 20 layers both architectures give similar results. However, when we stack more layers, the vanilla RHN stops improving (and even deteriorating), whereas RHN with HSG cell keeps improving. }
  \label{val_hist}
\end{figure}

\section{Conclusion}
In this work, we revisit a widely used RNN model. We analyze its limits and issues, and propose a variant for it called Highway State Gate (HSG). The main idea behind HSG is to generate an alternative fast route for the information to flow through time. The HSG uses a gating mechanism to assemble a new state out of a weighted sum of the former state and the RHN output. We show that when using our method, training deeper nets results in better performance. To the best of our knowledge, this is the first time in the field of Recurrent Nets that adding layers to this scale resulted in a steady improvement.

\bibliographystyle{plainnat}
\bibliography{ref}

\end{document}